\documentclass[10pt, conference, compsocconf]{IEEEtran}
\usepackage{cite}

\ifCLASSINFOpdf
 \usepackage{float}
  \usepackage[pdftex]{graphicx}
  \graphicspath{{images/}}
  \DeclareGraphicsExtensions{.pdf,.jpeg,.png}
\else
\fi
\usepackage{amsmath,amssymb,amstext}
\usepackage{physics}
\usepackage{balance}
\usepackage[linesnumbered,algoruled,boxed,lined]{algorithm2e}
\SetKwInOut{Parameter}{Parameters}

\usepackage{bm}

\usepackage{float}
\usepackage{booktabs, tabularx}

\newcolumntype{L}[1]{>{\raggedright\let\newline\\\arraybackslash\hspace{0pt}}m{#1}}
\newcolumntype{C}[1]{>{\centering\let\newline\\\arraybackslash\hspace{0pt}}m{#1}}
\newcolumntype{R}[1]{>{\raggedleft\let\newline\\\arraybackslash\hspace{0pt}}m{#1}}
\usepackage[subrefformat=parens,labelformat=parens]{subcaption}
\usepackage[export]{adjustbox} 

\usepackage{ifthen}
\newboolean{PrintVersion}
\setboolean{PrintVersion}{false}

\usepackage[pdftex,pagebackref=false]{hyperref} 
\hypersetup{
	plainpages=false,       
	unicode=false,          
	pdftoolbar=true,        
	pdfmenubar=true,        
	pdffitwindow=false,     
	pdfstartview={FitH},    
	pdftitle={Network Uncertainty Informed Semantic Feature Selection for Visual SLAM},    
	pdfauthor={Pranav Ganti, Steven L. Waslander},    
	pdfsubject={SLAM},  
	pdfnewwindow=true,      
	colorlinks=true,        
	linkcolor=blue,         
	citecolor=green,        
	filecolor=magenta,      
	urlcolor=cyan           
}
\ifthenelse{\boolean{PrintVersion}}{   
	\hypersetup{	
		citecolor=black,%
		filecolor=black,%
		linkcolor=black,%
		urlcolor=black}
}{} 

\begin{document}
	

\newcommand\scalemath[2]{\scalebox{#1}{\mbox{\ensuremath{\displaystyle #2}}}}

\newcommand{\state}{\mathbf{x}}
\newcommand{\roll}{\phi}
\newcommand{\pitch}{\psi}
\newcommand{\yaw}{\theta}

\newcommand{\GP}{\mathcal{GP}}
\newcommand{\normal}{\mathcal{N}}
\newcommand{\covar}{\mathbf{\Sigma}}
\newcommand{\mean}{\mu}
\newcommand{\meas}{\mathbf{z}}

\newcommand{\SO}{\mathbb{SO}(3)}
\newcommand{\so}{\mathfrak{so}(3)}
\newcommand{\SE}{\mathbb{SE}(3)}
\newcommand{\se}{\mathfrak{se}(3)}
\newcommand{\twist}{\bm{\xi}}
\newcommand{\rot}{\mathbf{R}}
\newcommand{\transf}{\mathbf{T}}
\newcommand{\trans}{\mathbf{t}}
\newcommand{\vel}{\mathbf{u}}
\newcommand{\point}{\mathbf{p}}
\newcommand{\soskew}{\bm{\omega}^{\wedge}}
\newcommand{\sorthree}{\bm{\omega}}
\newcommand{\semat}{\bm{\xi}^{\wedge}}

\newcommand{\K}{\mathbf{K}}
\newcommand{\I}{\mathbf{I}}

\newcommand{\cframe}{\mathcal{F}}

\newcommand{\adj}{\text{Adj}}

%
\title{Network Uncertainty Informed Semantic Feature Selection for Visual SLAM}


\author{\IEEEauthorblockN{Pranav Ganti}
\IEEEauthorblockA{Department of Mechanical and Mechatronics Engineering\\
University of Waterloo\\
Waterloo, Canada\\
pganti@uwaterloo.ca}
\and
\IEEEauthorblockN{Steven L. Waslander}
\IEEEauthorblockA{Institute for Aerospace Studies\\
University of Toronto\\
Toronto, Canada\\
stevenw@utias.utoronto.ca}
}


%


\maketitle

\begin{abstract}
In order to facilitate long-term localization using a visual simultaneous localization and mapping (SLAM) algorithm, careful feature selection can help ensure that reference points persist over long durations and the runtime and storage complexity of the algorithm remain consistent. We present SIVO (Semantically Informed Visual Odometry and Mapping), a novel information-theoretic feature selection method for visual SLAM which incorporates semantic segmentation and neural network uncertainty into the feature selection pipeline. Our algorithm selects points which provide the highest reduction in Shannon entropy between the entropy of the current state and the joint entropy of the state, given the addition of the new feature with the classification entropy of the feature from a Bayesian neural network. Each selected feature significantly reduces the uncertainty of the vehicle state and has been detected to be a static object (building, traffic sign, etc.) repeatedly with a high confidence. This selection strategy generates a sparse map which can facilitate long-term localization. The KITTI odometry dataset is used to evaluate our method, and we also compare our results against ORB\_SLAM2. Overall, SIVO performs comparably to the baseline method while reducing the map size by almost 70\%.

\end{abstract}

\begin{IEEEkeywords}
Localization, Mapping, SLAM, Deep Learning, Information Theory, Semantic Segmentation
\end{IEEEkeywords}

%
\IEEEpeerreviewmaketitle

\section{Introduction}
\label{sec:introduction}
Localization is a crucial problem for an autonomous vehicle. Accurate location knowledge facilitates a variety of tasks required for autonomous driving such as vehicle control, path planning or object tracking. Accurate positioning information is also a matter of safety, as localization accuracy must be known on the order of centimetres in order to prevent collisions and maintain lane positioning. Although sensors such as a Global Positioning System (GPS) can provide localization information to the desired accuracy, there are numerous situations where this is not possible, such passing through a tunnel or driving in dense urban environments. In recent years, visual odometry (VO)~\cite{nister2004visual} and visual simultaneous localization and mapping (SLAM) have emerged as reliable techniques for vehicle localization through the use of cameras. By observing the apparent motion of distinct reference points, or features, in the scene, we can determine the motion of a camera through the environment. The map generated by the SLAM algorithm can be used for long-term localization as it provides the vehicle with known references if it returns to a pre-mapped area. However, the runtime performance and storage requirements of the algorithm scale with the number of landmarks detected. Careful landmark selection can mitigate these issues as the robot navigates through its environment.

In order to accurately track camera motion, selected features should be: viewpoint, scale, rotation, illumination, and season invariant, as well as static. Traditional feature detectors and descriptors, such as SIFT~\cite{lowe2004distinctive}, SURF~\cite{bay2006surf}, FAST~\cite{rosten2006machine}, or ORB~\cite{rublee2011orb} aim to satisfy the first 3 criteria, while appearance based methods such as FAB-MAP~\cite{cummins2008fab} or SeqSLAM~\cite{milford2012seqslam} aim to fulfill criteria 4 and 5. Typically, visual SLAM algorithms depend on outlier rejection schemes such as RANSAC~\cite{fischler1981random} to characterize an object as dynamic. In this case, the motion of the dynamic reference point would be an outlier compared to the motion of static objects, which \emph{should} comprise the majority of the scene. This, however, is not always the case for autonomous driving due to other vehicles or pedestrians in the scene.
\begin{figure}[ht]
	\centering
	\includegraphics[width=0.48\textwidth]{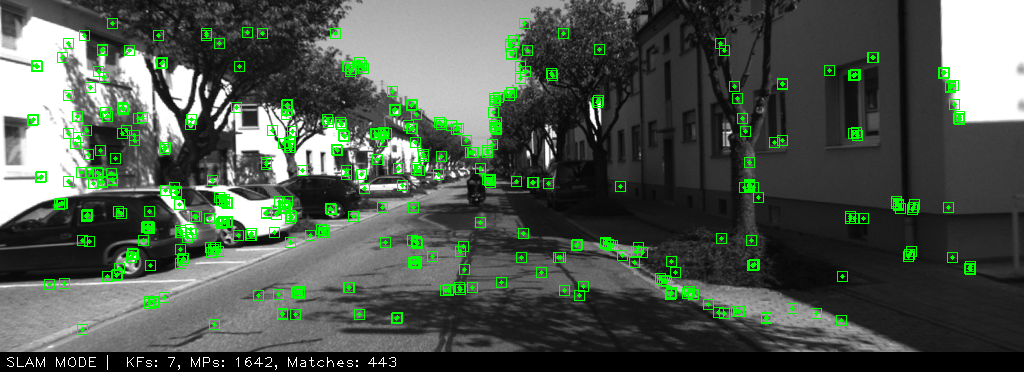}
	\setlength{\abovecaptionskip}{-10pt}
	\setlength{\belowcaptionskip}{-20pt}
	\caption{ORB\_SLAM2~\cite{mur2017orb} feature selection on KITTI~\cite{geiger2012kitti} sequence 00.}
	\label{fig:orb_features_00}
\end{figure}
\begin{figure}[ht]
	\includegraphics[width=0.48\textwidth]{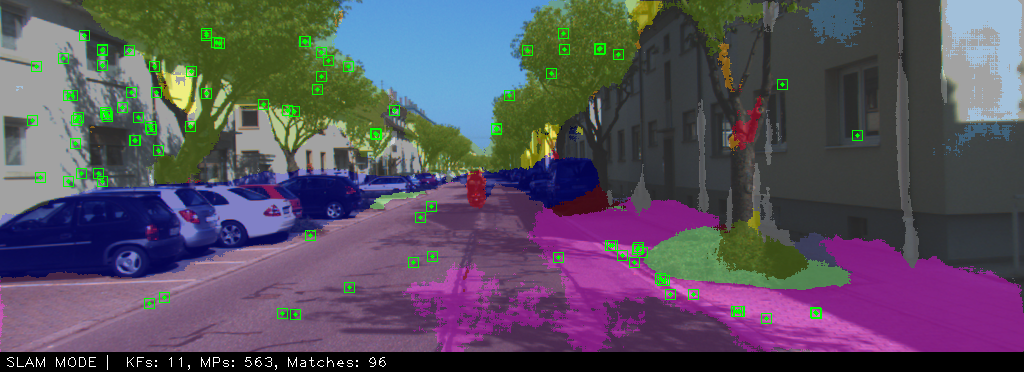}
	\setlength{\abovecaptionskip}{-10pt}
	\setlength{\belowcaptionskip}{-10pt}
	\caption{SIVO feature selection on KITTI sequence 00.}
	\label{fig:sivo_features_00}
\end{figure}
\begin{figure}[ht]
	\includegraphics[width=0.48\textwidth]{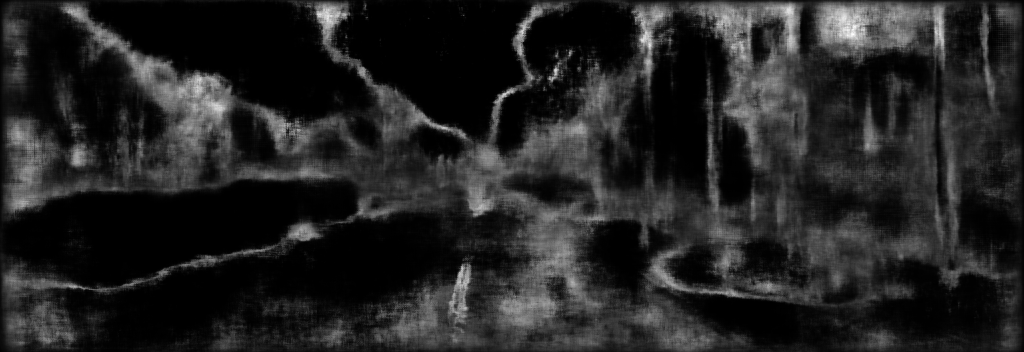}
	\setlength{\abovecaptionskip}{-10pt}
	\setlength{\belowcaptionskip}{-20pt}
	\caption{Variance image for Figure \ref{fig:sivo_features_00}. Black indicates low classification uncertainty, white indicates high uncertainty.}
	\label{fig:kitti_variance_00}
\end{figure}

Figure \ref{fig:orb_features_00} illustrates typical features used by a visual SLAM algorithm. The best reference points are most likely located on the buildings and signs. These are useful long-term references as they would only be modified in the event of construction or vandalism. In contrast, the vehicle features may be gone within the hour, and the foliage will no longer be present as the seasons change. The emergence of deep learning has led to rapid advances in scene understanding, allowing  context to be incorporated into SLAM and addressing our last criterion. We can now dictate which features are more likely to be stable from our contextual understanding of typical static and dynamic objects.

We present SIVO\footnote{Publicly available: \url{www.github.com/navganti/SIVO}} (Semantically Informed Visual Odometry and Mapping), a novel feature selection method for visual SLAM. This work enhances traditional feature detectors with deep learning based scene understanding using a Bayesian neural network (NN), which provides context for visual SLAM while accounting for neural network uncertainty. Our method selects features which provide the highest reduction in Shannon entropy between the entropy of the current state and the joint entropy of the state, given the addition of a new feature with the classification entropy of the feature from the Bayesian NN. This selection strategy generates a sparse map which can facilitate long-term localization, as each selected feature significantly reduces the uncertainty of the vehicle state and has been detected to be a static object (building, traffic sign, etc.) repeatedly with a high confidence. To the best of our knowledge, this is the first algorithm which directly considers NN uncertainty in an information-theoretic approach to SLAM.

\section{Related Works}
\label{sec:related_works}

Information-theoretic (IT) approaches have been prominent in maintaining the number of variables used by the SLAM optimization pipeline. These methods select features or keyframes which maximize the \emph{information gain} (see Section \ref{sec:background:information_theory}), and aim to reduce the number of optimization variables without appreciably compromising the accuracy of the SLAM solution. Dissanayake et al.~\cite{dissanayake2000computationally} propose a feature selection strategy which reduces the computational burden of maintaining a large map without affecting the statistical consistency of the estimation process. As the robot travels through its environment, a new landmark is only selected when the robot has travelled a predefined distance. The feature with the maximum information content is selected; this is determined by calculating the reciprocal of the trace of the landmark position covariance matrix. Hochdorfer and Schlegel~\cite{hochdorfer2009landmark} build on this method and consider spatial position in conjunction with landmark quality. Davison~\cite{davison2005active} proposes to use mutual information as the quality metric of a visual measurement. The landmark with the highest mutual information between itself and the robot pose will reduce pose uncertainty the most, and is selected to update the pose estimate. This process is repeated until the mutual information of the best feature drops below a predefined threshold. Kaess and Dellaert~\cite{kaess2009covariance} build on the method proposed by Davison and further save computing resources by immediately select all features which have a mutual information above a predefined threshold. Other information-theoretic methods include an entropy-based approach~\cite{zhang2005entropy}, an incremental approach which optimizes the tradeoff between memory consumption and estimation accuracy~\cite{choudhary2015information}, as well as an approach which uses Monte Carlo reinforcement learning~\cite{strasdat2009landmark} to learn the most useful landmarks. Map point maintenance is crucial for long-term localization, and information-theoretic approaches have proven to be effective in reducing the complexity of the SLAM problem. However, the maps generated by these algorithms will not have long-term stability due to the lack of semantic information.

The idea to incorporate semantic information into the SLAM pipeline is not novel. The emergence of deep learning has resulted in increasingly accurate methods to determine context within a scene. Salas-Moreno et al.~\cite{salas2013slam++} incorporate semantic information at the level of objects. In contrast to traditional SLAM algorithms, which use low-level primitive features such as points, lines, or patches for localization, the authors track the motion of objects in the scene from a known 3D object database. Murali et al.~\cite{murali2017utilizing} extend a custom map-builder and the localization functionality of ORB\_SLAM2~\cite{mur2017orb} to use semantic scene information obtained from a low-rank version of SegNet~\cite{badrinarayanan2017segnet}. A feature is deemed to be invalid if the class is a temporal object (car, bike, pedestrian), or too far away (sky, road). This feature selection scheme is only used for mapping, while all features are incorporated for visual odometry. Semantic information has also been included into direct methods~\cite{li2016semi}, and An et al.~\cite{an2017semantic} propose a VO pipeline which fuses semantics into a semi-direct method. Some semantic approaches bypass the use of feature detectors entirely. Stenborg et al.~\cite{stenborg2018long} use only the 3D location of a feature and its semantic label as a descriptor, and use a particle filter to bypass the use of traditional feature detectors. While the rapid advancement of deep learning has allowed for the development of semantic SLAM algorithms, most approaches to date treat network output as deterministic and do not account for uncertainty in the NN output. 

As we continue to develop our machine learning based methods, it is imperative to understand how much \emph{trust} we can place in the network and use this uncertainty to facilitate robot decisions. Mu et al.~\cite{mu2016slam} propose an object-based SLAM method which uses a nonparametric pose graph to couple together data association and SLAM. They model the NN object detection likelihood using a Dirichlet process and simultaneously optimize over the joint likelihood of the odometry, measurements, and class detections. Bowman et al.~\cite{bowman2017probabilistic} propose a similar pipeline which uses expectation maximization to jointly estimate data association and sensor states, however they specify their detection probability with covariance proportional to the size of the detected bounding box. 

Accounting for NN uncertainty in the SLAM pipeline is still an underexplored area. Gal~\cite{gal2016uncertainty, gal2015bayesian, gal2016dropout, gal2015dropout} and Kendall~\cite{kendall2017uncertainties, kendall2015bayesian, kendall2016modelling} have investigated the use of dropout~\cite{srivastava2014dropout} at test time to better estimate neural network uncertainty for both classification and regression tasks. The authors show that multiple forward passes with dropout is equivalent to approximating a Bayesian neural network, which allows for uncertainty to be extracted from the network output. Our approach uses this network uncertainty formulation in conjunction with a traditional IT approach, allowing us to reliably incorporate semantic information while maintaining map size.

\section{Problem Formulation}
\label{sec:background}

\subsection{Uncertainty Estimation for Semantic Segmentation}
\label{sec:background:uncertainty_ml}

In this work, we will be following the methodology presented by Gal~\cite{gal2016uncertainty, gal2015bayesian, gal2016dropout, gal2015dropout}.

\subsubsection{Bayesian Neural Networks}
\label{sec:background:gp}

It has been shown that an NN with one layer, an infinite number of weights, and a Gaussian distribution placed over each of its weights converges to a Gaussian Process (GP)~\cite{neal1995bayesian}. We can intuitively see that this is the case; NNs can be considered as ``function approximators", and placing a Gaussian distribution (typically a standard Gaussian) over the weights results in a distribution over the function. An infinitely-wide NN is obviously impossible to construct, however, finite NNs with distributions over its weights have been studied as \emph{Bayesian Neural Networks}~\cite{neal1995bayesian}. Gal~\cite{gal2016dropout} shows that applying dropout~\cite{srivastava2014dropout} before every weight layer in an NN with arbitrary depth and nonlinearities is a mathematically equivalent approximation to the Bayesian NN, which in turn is an approximation to the deep GP.

\subsubsection{Output Prediction using a Bayesian Neural Network}
\label{sec:output_pred}
Let us define the input to a neural network as $\mathbf{x} \in \mathbb{R}^{H\times W}$, and the weight matrix for each layer as $\mathbf{W}_i$, with $L$ total layers and varying dimensions. The output of an NN, $\mathbf{y} \in \mathbb{R}^C$ can be expressed by
\begin{equation}
\mathbf{y} = \bm{\mu}(\mathbf{x}, (\mathbf{W}_i)_{i=1}^L)
\end{equation}
where $\bm{\mu}: \mathbb{R}^{H\times W} \mapsto \mathbb{R}^C$ is the underlying function approximated by the Bayesian NN. The classification output is constructed by passing the network output through softmax function.
\begin{equation}
\label{eqn:softmax_bayesian_nn}
p(c|\mathbf{x}, (\mathbf{W}_i)_{i=1}^L) = \text{Categorical}\left(\frac{\exp(y_{c})}{\sum\limits_{c \in C} \exp(y_{c})}\right)
\end{equation}
where $p(c|\mathbf{x}, (\mathbf{W}_i)_{i=1}^L)$ represents the probability of a particular class output, $c$, out of $C$ possibilities, given the input and weights. For any new input $\mathbf{x^*} \in \mathbb{R}^{H\times W}$, the predicted output can be determined by integrating over all possible functions represented by the Bayesian NN. Let us define $\mathbf{X} \in \mathbb{R}^{H\times W\times Q}$ as our training data input, and $\mathbf{Y} \in \mathbb{R}^{C \times Q}$ as our training data output, indicating that we had $Q$ training examples. The probability for a new predicted output, $\mathbf{y^*} \in \mathbb{R}^C$ is defined by
\begin{equation}
\label{eqn:predictive_output_gaussian}
p(\mathbf{y}^*|\mathbf{x}^*, \mathbf{X}, \mathbf{Y}) = \int p(\mathbf{y}|\bm{\mu})p(\bm{\mu}|\mathbf{x}^*, \mathbf{X}, \mathbf{Y})d\bm{\mu}
\end{equation}
This integral is typically intractable, but can be approximated using variational inference and Monte Carlo integration~\cite{gal2016dropout}. In contrast to averaging the weights at test time as described in~\cite{srivastava2014dropout}, the same input is passed through the network repeatedly, and dropout is used at test time to provide a different ``thinned'' network for each trial. The outputs from each trial are then averaged to provide the final output. This is referred to as \emph{MC (Monte Carlo) dropout}~\cite{gal2015bayesian}.
\begin{equation}
\label{eqn:mc_dropout}
p(\mathbf{y}^*|\mathbf{x}^*, \mathbf{X}, \mathbf{Y}) \approx \frac{1}{N} \sum\limits_{n=1}^Np(\mathbf{y}^*|\mathbf{x}^*, \mathbf{W})
\end{equation}
where $N$ is the number of passes through the network and $\mathbf{W}$ represents some non-zero subset of the weights after applying dropout. It is important to note the distinction between the softmax output and the result of Monte Carlo sampling; the softmax mapping describes relative probabilities between class detections, but is not an absolute measure of uncertainty.

\subsection{Information Theory}
\label{sec:background:information_theory}

For a stochastic variable $\mathbf{X} = \left\{x_0, x_1, \dots, x_k\right\} \in \mathbb{R}^K$ with probability mass function $p(x)$, the entropy, or average uncertainty, is defined by
\begin{equation}
\label{eqn:entropy}
H(x) = -\sum\limits_{x \in \mathbf{X}} p(x)\log p(x)
\end{equation}
and is measured in \emph{bits}. The entropy of a multivariate Gaussian variable, $\mathbf{x}$, is defined by
\begin{equation}
\label{eqn:entropy_multivariate_gaussian}
H(\state) = \frac{1}{2}\log((2\pi e)^n\det(\covar))
\end{equation}
where $\covar$ represents the covariance matrix, and $n$ is the dimension of the random variable.
Let us define two dependent random variables, $\mathbf{X} = \left\{x_0, x_1, \dots, x_k\right\} \in \mathbb{R}^K$ and $\mathbf{Y}  = \left\{y_0, y_1, \dots, y_v\right\} \in \mathbb{R}^V$ with pdfs $p(x)$ and $p(y)$ respectively. The mutual information, or information \emph{shared} between the two variables is represented by~\cite{cover2012elements}
\begin{equation}
\label{eqn:mutual_information}
I(\mathbf{X}; \mathbf{Y}) = \sum_{x \in \mathbf{X}} \sum_{y \in \mathbf{Y}} p(x, y)\log \frac{p(x, y)}{p(x)p(y)}
\end{equation}
The mutual information between two parts of a multivariate Gaussian, $\state$, is defined by~\cite{chli2010applying}
\begin{equation}
\state = \begin{bmatrix}
\mathbf{a}\\
\mathbf{b}
\end{bmatrix},
\quad
\covar = \begin{bmatrix}
\covar_{aa} & \covar_{ab}\\
\covar_{ba} & \covar_{bb}
\end{bmatrix}
\end{equation}
\begin{equation}
\label{eqn:MI}
I(\mathbf{a};\mathbf{b}) = \frac{1}{2}\log\left(\frac{\det(\covar_{aa}) \det(\covar_{bb})}{\det(\covar)}\right)
\end{equation}
Mutual information and entropy are tightly coupled. The mutual information between variables $\mathbf{X}$ and $\mathbf{Y}$ can also be represented by~\cite{cover2012elements}
\begin{equation}
\label{eqn:mi_entropy}
I(\mathbf{X};\mathbf{Y}) = H(\mathbf{X}) - H(\mathbf{X}|\mathbf{Y})
\end{equation}

\subsection{Uncertainty in Classification Results for a Bayesian Neural Network}
\label{sec:background:uncertainty_ml_classification}

Entropy can be used as a metric for classification uncertainty from a Bayesian NN~\cite{gal2016uncertainty}. Let us denote  $\mathbf{y} \in \mathbb{R}^C$ as our network output, $\bm{\mathcal{I}} \in \mathbb{R}^{W \times H}$ as our input image data, $\bm{\mathcal{D}} \in \mathbb{R}^{W \times H \times Q}$ as our training data, and $c$ as a particular class output with $c \in C$ potential classes. The entropy is defined by~\cite{gal2016uncertainty}
\begin{equation}
\label{eqn:entropy_classification}
H(c| \bm{\mathcal{I}}, \bm{\mathcal{D}}) := - \sum_{c \in C} p(c|\bm{\mathcal{I}}, \bm{\mathcal{D}})\log p(c|\bm{\mathcal{I}}, \bm{\mathcal{D}})
\end{equation}
Equation \ref{eqn:entropy_classification} reaches a maximum value when all of the class outputs are equiprobable, and a minimum value of $0$ when one class is predicted with a probability of $1$. Although the individual confidence values do not have any meaning of uncertainty, the entropy calculation will observe the spread in the confidence value for each class output of a pixel. We can write an expression for the approximate entropy for the confidence output in bits by substituting Equation \ref{eqn:mc_dropout} into Equation \ref{eqn:entropy_classification}~\cite{gal2016uncertainty}.
\begin{equation}
\label{eqn:entropy_classification_approx}
\scalemath{0.85}{H(c|\bm{\mathcal{I}}, \bm{\mathcal{D}}) = -\sum_{c\in C}\left(\frac{1}{N}\sum_n^N p(c| \bm{\mathcal{I}}, \mathbf{W})\right) \log\left(\frac{1}{N}\sum_n^N p(c | \bm{\mathcal{I}}, \mathbf{W})\right)}
\end{equation}
%
\section{Feature Selection Criteria}
\label{sec:problem_formulation}

We now present our feature selection methodology for semantic visual SLAM with neural network uncertainty. Our method builds upon the work of Davison~\cite{davison2005active} and the enhancement by Kaess and Dellaert~\cite{kaess2009covariance}. We will first outline these methods in detail, and then present SIVO.

\subsection{Information-Theoretic Feature Selection Criteria}

Let us denote the 6DOF state parameterization at some time $t$ as $\state_t \in \SE$. As our main goal is to track camera poses through time, the state represents the \emph{pose} of the camera frame $c$ with respect to the \emph{world} frame $w$ at time $t$. This can also be represented by $\transf_{cw}^t \in \SE$ with associated covariance matrix $\covar_t \in \mathbb{R}^{6 \times 6}$. Measurements are defined through a nonlinear measurement model, $h(\state_t)$, as
\begin{equation}
\meas_i = h_i(\state_t) + \bm{\epsilon}, \quad \bm{\epsilon} \sim \normal (\mathbf{0}, \mathbf{Q}_i)
\end{equation}
where $\meas_i$ represents the feature measurement, and $\epsilon$ is zero-mean Gaussian noise with noise covariance $\mathbf{Q}_i \in \mathbb{R}^{3 \times 3}$. The measurement model is the rectified stereo projection model, $\pi_s : \mathbb{R}^3 \mapsto \mathbb{R}^3$, where we assume that the transformation between the right and left cameras is a purely horizontal translation equivalent to the baseline. We define ${_c}x, {_c}y, {_c}z$ as the $x$, $y$, and $z$-coordinates of the point in the camera frame ${_c}\point$. The point in the world frame is defined as ${_w}\point$, the camera intrinsic parameters are $f_x, f_y, c_x, c_y$, and $b$ as the baseline between stereo cameras.
\begin{equation*}
\begingroup
\renewcommand*{\arraystretch}{1.5}
h_i (\state_t) = \pi_s (\transf_{cw}^t {_w}\point) = \pi_s(_c\point)
= \pi_s(_cx \quad _cy \quad _cz)^T
\endgroup
\end{equation*}
\begin{equation}
\label{eqn:stereo_proj}
\begingroup
\renewcommand*{\arraystretch}{1.5}
h_i(\state_t) =
\begin{bmatrix}
f_x\frac{_cx}{_cz} + c_x\\
f_y\frac{_cy}{_cz} + c_y\\
f_x\frac{(_cx - b)}{_cz} + c_x
\end{bmatrix}
\endgroup
\end{equation}
Assume that at some  time $t$, we have $n$ available features distributed throughout the scene that we can select for our SLAM algorithm. We can stack the current pose with the candidate measurements into a vector as such
\begin{equation}
\hat{\state} =
\begin{bmatrix}
\state_t \quad h_1(\state) \quad h_2(\state) \quad \hdots \quad h_n(\state)
\end{bmatrix}^T
\end{equation}
As each of these random variables are described by multivariate Gaussians, it follows that the stacked vector, $\hat{\state}$, is also a multivariate Gaussian. This stacked variable also has a covariance matrix which consists of the pose covariance and measurement covariances, where the latter is calculated by propagating the state covariance through the measurement model. This is defined by
\begin{equation}
\label{eqn:lifted_covar}
\begingroup
\renewcommand*{\arraystretch}{1.5}
\hat{\bm{\Sigma}} =
\scalemath{0.835925}{\begin{bmatrix}
	\covar_t & \covar_t\frac{\partial h_1}{\partial \state}^T & \cdots & \covar_t \frac{\partial h_n}{\partial \state}^T\\
	\frac{\partial h_1}{\partial \state}\covar_t & \frac{\partial h_1}{\partial \state}\covar_t\frac{\partial h_1}{\partial \state}^T + \mathbf{Q}_1 & \cdots & \frac{\partial h_1}{\partial \state}\covar_t\frac{\partial h_n}{\partial \state}^T \\
	\vdots & \vdots & \ddots & \vdots\\
	\frac{\partial h_n}{\partial \state}\covar_t & \frac{\partial h_n}{\partial \state}\covar_t\frac{\partial h_1}{\partial \state}^T & \cdots & \frac{\partial h_n}{\partial \state}\covar_t\frac{\partial h_n}{\partial \state}^T + \mathbf{Q}_n
	\end{bmatrix}}
\endgroup
\end{equation}
The measurement which best reduces the pose uncertainty is then selected to update the state. This measurement has the maximum mutual information between the state and measurement, and can be easily calculated using Equation \ref{eqn:MI}. However, the \emph{marginal covariance} for each feature $\meas_i$, must first be constructed by selecting the relevant variables from Equation \ref{eqn:lifted_covar}.
\begin{equation}
\label{eqn:marg_covar}
\begingroup
\renewcommand*{\arraystretch}{1.5}
\hat{\covar}_i =
\begin{bmatrix}
\covar_{\state\state} & \covar_{\state\meas_i}\\
\covar_{\meas_i\state} & \covar_{\meas_i\meas_i}
\end{bmatrix} =
\begin{bmatrix}
\covar_t & \covar_t\frac{\partial h_i}{\partial \state}^T\\
\frac{\partial h_i}{\partial \state}\covar_t & \frac{\partial h_i}{\partial \state}\covar_t\frac{\partial h_i}{\partial \state}^T + \mathbf{Q}_i
\end{bmatrix}
\endgroup
\end{equation}
Once the state has been updated, this process is repeated until the maximum information provided by a new measurement falls below a user-defined threshold value.

Kaess and Dellaert~\cite{kaess2009covariance} build on Davison's method. The authors argue that selecting individual features and then updating the state estimate is not practical, as updating the state and extracting the marginal pose covariance values prior to taking each measurement can be quite expensive. Therefore, all measurements which have a mutual information above a predefined threshold are selected, and only then is the pose estimate updated. Although this is will not guarantee that the optimal landmark is selected, it is less computationally expensive. This approach forms the foundation for SIVO.

\subsection{SIVO Feature Selection Criteria}

The information-theoretic approach is now modified to incorporate semantic information. Each measurement from Equation \ref{eqn:stereo_proj} is a stereo projection of a 3D point into the image space. Using semantic segmentation, a discrete class value for each pixel can also be determined, providing context to the measurement.

Using Equation \ref{eqn:mi_entropy}, the mutual information-based criteria from Equation \ref{eqn:MI} can be expressed in terms of entropy.
\begin{equation}
\label{eqn:dH}
I(\state; \meas_i) =  \Delta H_i = H(\state | \mathbf{Z}) - H(\state | \meas_i, \mathbf{Z})
\end{equation}
where $\mathbf{Z}$ represents all \emph{previous} measurements made in order to obtain our current state estimate. If $\Delta H_i$ is greater than a predefined threshold for a measurement, $\meas_i$, it is selected as a reference for the SLAM algorithm.

We propose to modify $\Delta H_i$, and evaluate the entropy difference between the current state and the joint entropy of the state given the new feature measurement and the semantic segmentation classification, using Equation \ref{eqn:entropy_classification_approx}. This can be expressed as
\begin{equation}
\label{eqn:sivo_deltaH}
\Delta H = H(\state | \mathbf{Z}) - H(\state, c_i | \meas_i, \mathbf{Z}, \bm{\mathcal{I}}, \bm{\mathcal{D}})
\end{equation}
where $\bm{\mathcal{I}}$ represents the current image and $\bm{\mathcal{D}}$ represents the dataset used to train the neural network. We assume that the classification entropy and state entropy are conditionally independent, and therefore express the latter term in Equation \ref{eqn:sivo_deltaH} as
\begin{equation}
\begin{aligned}
H(\state, c_i | \meas_i, \mathbf{Z}, \bm{\mathcal{I}}, \bm{\mathcal{D}}) &= H(\state | \meas_i, \mathbf{Z}, \bm{\mathcal{I}}, \bm{\mathcal{D}}) + H(c_i | \meas_i, \mathbf{Z}, \bm{\mathcal{I}}, \bm{\mathcal{D}})\\
&= H(\state | \meas_i, \mathbf{Z}) + H(c_i | \bm{\mathcal{I}}, \bm{\mathcal{D}})
\end{aligned}
\end{equation}
The state is not dependent on the actual image or dataset, thus the conditionally dependent  terms can be removed from the individual entropy terms. Similarly, the classification detection is not dependent on any of the feature measurements. Therefore, Equation \ref{eqn:sivo_deltaH} can be rewritten as
\begin{equation}
\label{eqn:dh_intermed}
\Delta H_i = H(\state | \mathbf{Z}) -  H(\state | \meas_i, \mathbf{Z}) - H(c_i | \bm{\mathcal{I}}, \bm{\mathcal{D}})
\end{equation}
The first two terms are exactly the mutual information criterion from Equation \ref{eqn:dH}. Substituting Equation \ref{eqn:dH} into Equation \ref{eqn:dh_intermed} yields the SIVO feature selection threshold.
\begin{equation}
\label{eqn:sivo_selection_criteria}
\boxed{\Delta H_i = I(\state; \meas_i) - H(c_i | \bm{\mathcal{I}}, \bm{\mathcal{D}})}
\end{equation}

We argue that the best reference points should not only provide the most information to reduce the uncertainty of the state, but they should be static reference points which have been detected as such with a very high certainty. This feature selection criteria allows us to balance the value of a feature for the state estimate and the certainty of the feature's classification. Recall from Section \ref{sec:background:uncertainty_ml_classification} that the minimum value of the classification entropy is $0$ when the network predicts one class with a confidence of $100\%$. Therefore, in an ideal world where the class of each pixel is perfectly identified, features will be selected according to the original mutual information based criterion from Equation \ref{eqn:MI}, as long as they have been classified as static.

\section{Experimental Validation}
\label{sec:experimental_validation}

\subsection{Implementation and Training}

The localization functionality of SIVO is built on top of ORB\_SLAM2~\cite{mur2017orb}, and all loop closure and relocalization functionality is enabled. Bayesian SegNet~\cite{kendall2015bayesian} is used to semantically segment the images and provide network uncertainty. Network inference is implemented using Caffe's~\cite{jia2014caffe} C++ interface in order to integrate the results from Bayesian SegNet with ORB\_SLAM2. SIVO is publicly available on Github\footnote{\url{https://www.github.com/navganti/SIVO}}, and the training setup can also be found online\footnote{\url{https://www.github.com/navganti/SegNet}}.

Bayesian SegNet is trained using the Cityscapes dataset~\cite{cordts2016cityscapes} and then fine-tuned using the KITTI semantic~\cite{geiger2012kitti} dataset. The network was trained to segment 15 classes, where road, sidewalk, building, wall/fence, pole, traffic light, traffic sign, vegetation, and terrain are considered static, while sky, person/rider, car, truck/bus, motorcycle/bicycle, and void classes are considered to be dynamic, and are therefore ignored. We maintain the MC dropout probability of 50\% used in the Bayesian SegNet paper, and use the ``basic'' configuration of Bayesian SegNet in order to preserve GPU memory and speed up inference time. This architecture contains fewer layers in both the encoder and decoder compared to the original network.

\subsection{Results}

The KITTI~\cite{geiger2012kitti} odometry dataset is used to validate the performance of SIVO. The tunable parameters are the feature selection entropy threshold ($\Delta H$) and the number of samples for MC Dropout ($N$).The experiments will be referred to as follows: \textbf{B}ayesian \textbf{S}egNet $N$ \textbf{E}ntropy $\Delta H$. For example, an experiment where $N = 6$, and $\Delta H$ is set to $4$ bits is denoted as \textbf{BS6E4}. The following configurations are evaluated: BS2E4, BS6E2, BS6E3, BS6E4, and BS12E4. These experimental parameters are guided by the previous works. Kaess and Dellaert \cite{kaess2009covariance} ignore all features that have a mutual information below 2 bits, and Kendall et al. \cite{kendall2015bayesian} show that MC dropout outperforms the traditional weight averaging technique after approximately 6 samples. 

The same metrics used by the KITTI odometry benchmark are used to compare SIVO results to the KITTI ground truth and ORB\_SLAM2. First, both rotation and translational errors for all \emph{subsequences} from lengths 100m to 800m are evaluated. These values are then averaged over the subsequence lengths to provide a final translation error (\%) and rotation error (deg/m) for each trajectory. Table \ref{tbl:results} contains the compiled results for all trajectory results as well as the number of map points used by the algorithms, and is organized by translation error performance compared to ORB\_SLAM2.
\begin{table*}[ht]
	\caption{Translation error, rotation error, and map reduction for ORB\_SLAM2 and SIVO on the KITTI dataset.}
	\setlength{\belowcaptionskip}{-15pt}
	\label{tbl:results}
	\centering
	\begin{tabular}{C{1cm} | C{1.65cm} C{1.5cm} C{1.65cm} C{1.5cm} C{1.5cm} C{1.5cm} C{1.9cm} C{1cm}}
		\toprule
		\textbf{KITTI} \newline \textbf{Sequence} & \textbf{ORB Trans. Err. (\%)} & \textbf{ORB Rot. Err. (deg/m)} & \textbf{SIVO Trans. Err. (\%)} & \textbf{SIVO Rot. Err. (deg/m)} & \textbf{ORB Map Points} & \textbf{SIVO Map Points} & \textbf{Map} \newline \textbf{ Reduction (\%)} &\textbf{SIVO} \newline \textbf{Config.} \\
		\midrule
		09 & 1.20 & $4.63\mathrm{E}{-5}$ & $\mathbf{1.18}$ & $\mathbf{3.60\bm{\mathrm{E}}{-5}}$ & 64,442 & \textbf{18,893} & 70.68 & BS6E2\\
		10 & $\mathbf{0.95}$ & $\mathbf{4.85\bm{\mathrm{E}}{-5}}$ & 0.97 & $7.34\mathrm{E}{-5}$ & 33,181 & \textbf{9,369} & 71.76 & BS2E4\\
		08 & $\mathbf{1.15}$ & $\mathbf{4.89\bm{\mathrm{E}}{-5}}$ & 1.29 & $4.98\mathrm{E}{-5}$ & 127,810 & \textbf{40,574} & 68.25 & BS6E3\\
		05 & $\mathbf{0.59}$ & $\mathbf{2.70\bm{\mathrm{E}}{-5}}$ & 0.76 & $2.93\mathrm{E}{-5}$ & 73,463 & \textbf{22,237} & 69.73 & BS6E3\\	
		07 & $\mathbf{0.58}$ & $\mathbf{4.43\bm{\mathrm{E}}{-5}}$ & 0.80 & $5.08\mathrm{E}{-5}$ & 29,632 & \textbf{9,684} & 67.32 & BS6E3\\
		00 & $\mathbf{1.18}$ & $\mathbf{3.88\bm{\mathrm{E}}{-5}}$ & 1.44 & $4.68\mathrm{E}{-5}$ & 138,153 & \textbf{45,875} & 66.79 & BS6E4\\
		02 & $\mathbf{1.37}$ & $\mathbf{3.95\bm{\mathrm{E}}{-5}}$ & 1.70 & $4.86\mathrm{E}{-5}$ & 202,293 & \textbf{58,894} & 70.89 & BS12E4\\
		\midrule
		04 & $\mathbf{0.67}$ & $2.20\mathrm{E}{-5}$ & 1.50 & $\mathbf{1.97\bm{\mathrm{E}}{-5}}$ & 21,056 & \textbf{6,328} & 69.95 & BS12E4\\
		03 & $\mathbf{2.72}$ & $\mathbf{3.75\bm{\mathrm{E}}{-5}}$ & 4.65 & $1.29\mathrm{E}{-4}$& 27,209 & \textbf{8,449} & 68.95 & BS12E4\\
		01 & $\mathbf{1.01}$ & $\mathbf{3.10\bm{\mathrm{E}}{-5}}$ & 3.17 & $9.31\mathrm{E}{-5}$ & 101,378 & \textbf{37,233} & 63.27 & BS2E4\\
		06 & $\mathbf{0.67}$ & $\mathbf{3.91\bm{\mathrm{E}}{-5}}$ & 7.10 & $3.27\mathrm{E}{-4}$ & 47,461 & \textbf{11,396} & 75.99 & BS6E3\\
		\midrule
		\textbf{Average} & $\mathbf{1.10}$ & $\mathbf{3.84\bm{\mathrm{E}}{-5}}$ & 2.23 & $8.21\mathrm{E}{-5}$ & - & - & 69.42 & -\\
		\bottomrule
	\end{tabular}
	\vspace{-5.0mm}
\end{table*}
In summary, SIVO outperformed ORB\_SLAM2 on KITTI sequence 09 (Figure \ref{fig:kitti_09}), and performed comparably albeit less accurate (average of 1.13\% translation error difference, $4.37 \times 10^{-5}$ deg/m rotation error difference) while removing 69.4\% of the map points on average. This comparable performance indicates that the points removed by SIVO are redundant and the remaining points should be excellent long-term reference points for visual SLAM, although it is not possible to verify feature persistence with the KITTI dataset.
\begin{figure}[ht]
	\centering
	\includegraphics[width=0.48\textwidth]{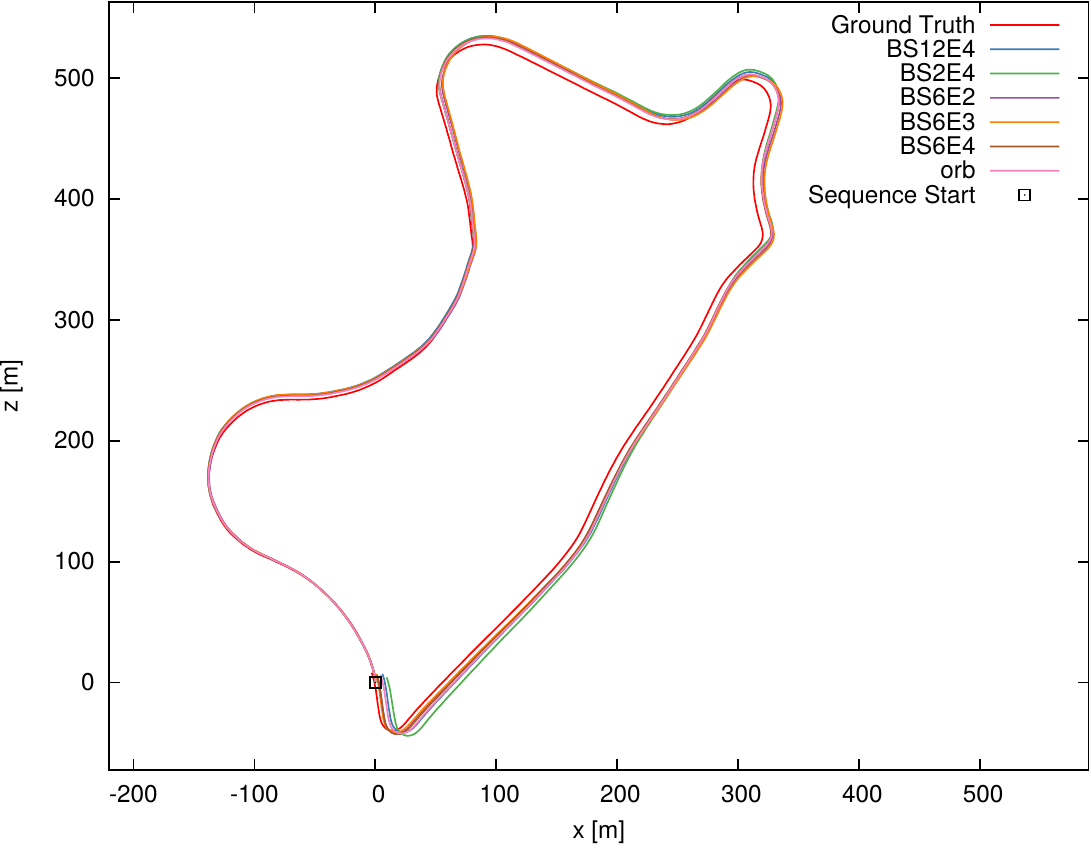}
	\setlength{\abovecaptionskip}{-10pt}
	\setlength{\belowcaptionskip}{-15pt}
	\caption{Trajectory overlay of SIVO, ORB\_SLAM2, and KITTI ground truth on sequence 09. SIVO has the lowest translation and rotation error on this sequence.}
	\label{fig:kitti_09}
\end{figure}

\section{Discussion}

Overall, our feature selection scheme is a good first approach to incorporating neural network uncertainty into a visual SLAM formulation. The results between SIVO and ORB\_SLAM2 are comparable, with a translation error difference of 1.13\% and a rotation error difference of $4.37\times10^{-5}$deg/m even when removing an average of 69.4\% of the map points used by the optimization pipeline. SIVO successfully removed points from the environment which are uninformative and/or dynamic. Figure \ref{fig:sivo_features_00} illustrates features selected on KITTI sequence 00, while Figure \ref{fig:kitti_variance_00} shows the \emph{variance} image for the scene. The variance image shows the spread of classification confidence from the trials of MC dropout, discussed in Section \ref{sec:output_pred}; black indicates a normalized variance of 0, while white indicates a normalized variance of 1. SIVO has mostly selected features which have a low variance, however the occasional uncertain point (such as the windowsill on the right side) has been selected as it sufficiently reduces the pose uncertainty.

In some cases, however, removing these map points did have an adverse effect on localization performance, which can be mostly attributed to the removal of short range feature points. SIVO immediately removes a point if it has been designated as a dynamic class, however, the KITTI odometry set has been curated to contain mostly static scenes and most sequences contain numerous parked cars. This difference is illustrated in Figures \ref{fig:orb_features_00} and \ref{fig:sivo_features_00}. These cars make up the majority of close features in the scene, which is required to better estimate translation. To accurately estimate camera pose, an even distribution of points throughout the image and 3D space is required. Far points will help with rotation estimation but are poor translation estimates, and close points can help with both. For all trajectories, ORB\_SLAM2 and SIVO have comparable, accurate rotation estimation, but SIVO generally performs worse in estimating translation. The four trajectories which performed significantly worse (sequences 01, 03, 04, and 06 in Table \ref{tbl:results}, separated by the midline) are all ``straight-line'' trajectories. In these sequences, the apparent motion of the features is quite small, as they are far away and lie directly ahead of the vehicle; this results in significant translation error. 

\begin{figure}[ht]
	\centering
	\includegraphics[width=0.48\textwidth]{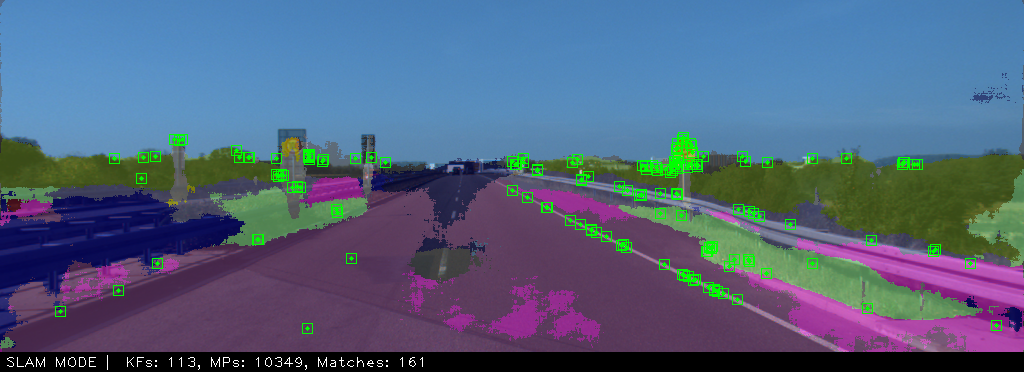}
	\setlength{\abovecaptionskip}{-10pt}
	\setlength{\belowcaptionskip}{-10pt}
	\caption{Low quality segmentation on KITTI sequence 01.}
	\label{fig:kitti_01_poor_seg}
	\vspace{-0.5mm}
\end{figure}

The localization performance is also dependent on segmentation quality. For example, sequence 01 (the ``highway'' sequence), in addition to being a mostly ``straight-line'' sequence, is not well represented in the semantic dataset. In Figure \ref{fig:kitti_01_poor_seg}, part of the highway divider as well as the bridge in the distance are misclassified as a car, which is immediately ignored by the algorithm. These features make up most of the close features for this sequence; SIVO is therefore relying on further features for this trajectory, and the translation performance suffers as a result.

A notable observation from Figures \ref{fig:orb_features_00} and \ref{fig:sivo_features_00} is the distribution of features throughout the scene. The ORB features extracted from the environment are corners, which will typically lie at the border between objects. However, as seen in Figure \ref{fig:kitti_variance_00}, these edges are where the neural network has the least confidence in its segmentation result. This behaviour is expected. The network struggles to generalize to border cases as they differ drastically between training examples. SIVO will consider a feature's value through its mutual information, but several of the strongest ORB candidates are eliminated as they have a higher entropy. Removing some of these features does have a benefit; although they have a strong response, they are generally composed of pixels from significantly different depths, resulting in inaccurate triangulation and poor tracking across multiple viewpoints. 

\section{Conclusion}
\label{sec:conclusion}

We present SIVO (Semantically Informed Visual Odometry and Mapping), a novel feature selection algorithm for visual SLAM which fuses together NN uncertainty with an information-theoretic approach to visual SLAM. SIVO outperformed ORB\_SLAM2 on KITTI sequence 09, and performed comparably well on 6 of the 10 remaining trajectories while removing 69.4\% of the map points on average. While incorporating semantic information into the SLAM algorithm is not novel, to the best of our knowledge this is the first algorithm which directly considers NN uncertainty in an information-theoretic approach to SLAM. Our method selects points which significantly reduce the Shannon entropy between the current state entropy and the joint entropy of the state, given the addition of the new feature with the classification entropy of the feature from the Bayesian NN. Each selected feature significantly reduces the uncertainty of the vehicle state and has been detected to be a static object repeatedly with a high confidence, resulting in a sparse map which can facilitate long-term localization. Our future work aims to refine segmentation performance and verify long-term localization capability on different datasets. We also will look to introduce further context and determine whether an observed vehicle is static or dynamic. This would allow for the use of short range features detected on static vehicles in a visual odometry solution for local pose estimation, while still ignoring these points in map creation.



%
\balance
\bibliographystyle{IEEEtran}
\bibliography{bib/bibliography}

\end{document}